\documentclass[10pt, conference, compsocconf]{IEEEtran}
\usepackage{times}
\usepackage{epsfig}
\usepackage{graphicx}
\usepackage{amsmath}
\usepackage{amssymb}
\newcommand{\tabincell}[2]{\begin{tabular}{@{}#1@{}}#2\end{tabular}} 
\usepackage[pagebackref=true,breaklinks=true,letterpaper=true,colorlinks,bookmarks=false]{hyperref}

\begin{document}
%
\title{Template-Instance Loss for Offline Handwritten Chinese Character Recognition}

\author{\IEEEauthorblockN{Yao Xiao$^1$, Dan Meng$^1$, Cewu Lu$^2$, Chi-Keung Tang$^3$}
\IEEEauthorblockA{$^1$Institute of Artificial Intelligence, Shanghai Em-Data Technology Co., Ltd., Shanghai, China\\
$^2$Shanghai Jiao Tong University, Shanghai, China\\
$^3$Hong Kong University of Science and Technology, Hong Kong, China\\
Email: \{xiaoyao,daisymeng\}@em-data.com.cn, lucewu@sjtu.edu.cn, cktang@cs.ust.hk}
}

\maketitle

\begin{abstract}
   The long-standing challenges for offline handwritten Chinese character
recognition (HCCR) are twofold: Chinese characters can be very diverse and
complicated while similarly looking, and cursive handwriting (due to
increased writing speed and infrequent pen lifting) makes strokes and even
characters connected together in a flowing manner. In this paper, we
propose the template and instance loss functions for the relevant machine 
learning tasks in offline handwritten Chinese character recognition.  
First, the character template is designed to deal with the intrinsic similarities 
among Chinese characters.  
Second, the instance loss can reduce category variance according to classification difficulty,
giving a large penalty to the outlier instance of handwritten Chinese character.
Trained with the new loss functions using our deep network architecture 
HCCR14Layer model consisting of simple layers, our extensive experiments show 
that it yields state-of-the-art performance and beyond for offline HCCR.

\end{abstract}


%
\IEEEpeerreviewmaketitle

\section{Introduction}

HCCR has been extensively studied 
for over 40 years. To deal with the challenge of large number of character classes, numerous
approaches have been proposed to improve recognition accuracy, though the recognition
results were still unsatisfactory. In hindsight, the recognition task is extremely 
challenging due to two reasons: 1) many Chinese characters have intrinsic similarities 
in structure; 2) great diversity of handwriting style makes the recognition problem 
more difficult.  Traditional handcrafted features are incapable of representing well 
the inherent high complexity of Chinese characters. We believe that the limit has reached for 
handcrafted features for handwritten Chinese characters recognition, 
but unfortunately the pertinent recognition accuracy is still far from human performance.

\begin{figure}[t]
	\centering
	\begin{tabular}{@{\hspace{0mm}}c@{\hspace{1mm}}c@{\hspace{1mm}}c}
		\includegraphics[width=0.32\linewidth]{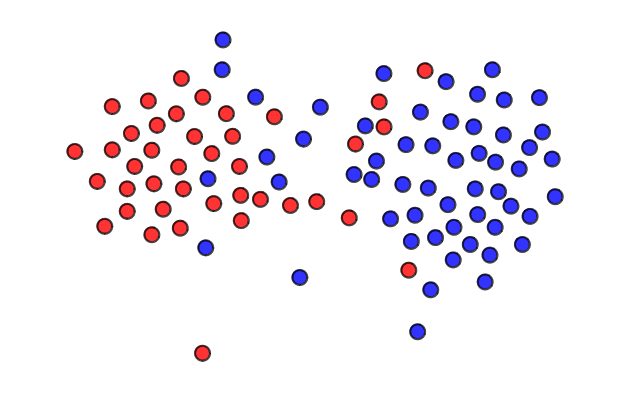} &
		\includegraphics[width=0.32\linewidth]{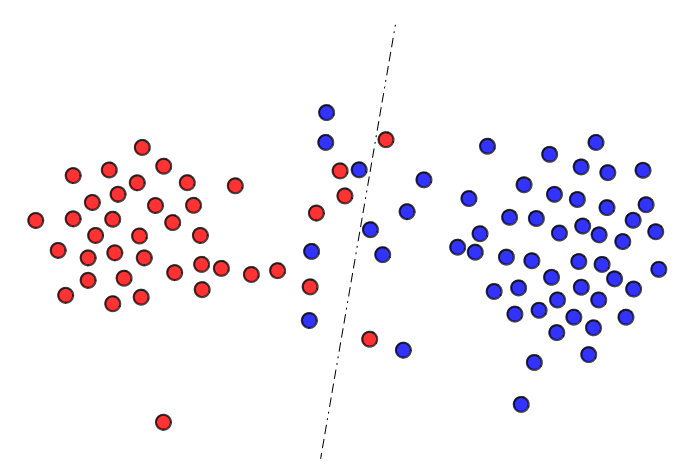} &
		\includegraphics[width=0.32\linewidth]{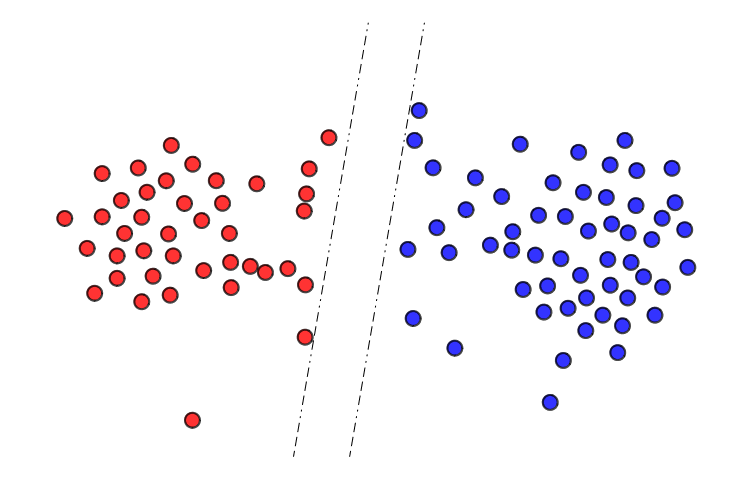} \\
		& & \\
		(a) Original & (b) Template & (c) Template+instance
	\end{tabular}
	\caption{Illustration of template-instance loss. The template loss works on category 
		level, and the instance loss affects misclassified samples.}
	\label{fig:teaser1}
\end{figure}

\begin{figure*}[t]
	\centering
	\begin{tabular}{@{\hspace{0mm}}c@{\hspace{1mm}}c@{\hspace{1mm}}c}
		\includegraphics[width=0.27\linewidth]{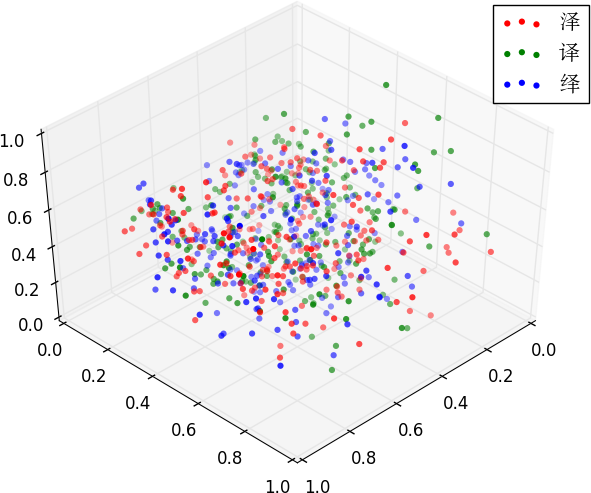} &
		\includegraphics[width=0.27\linewidth]{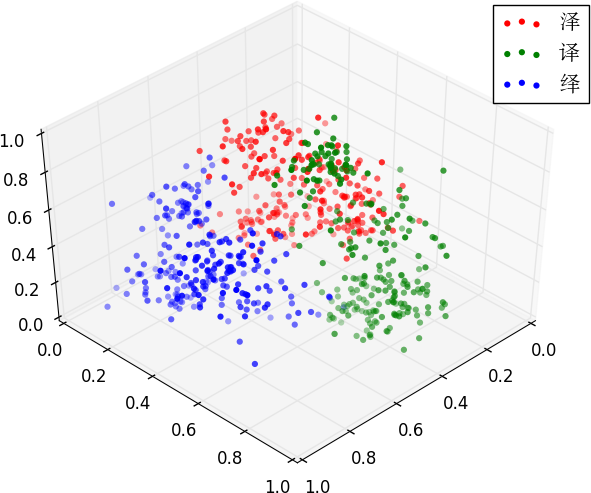} &
		\includegraphics[width=0.27\linewidth]{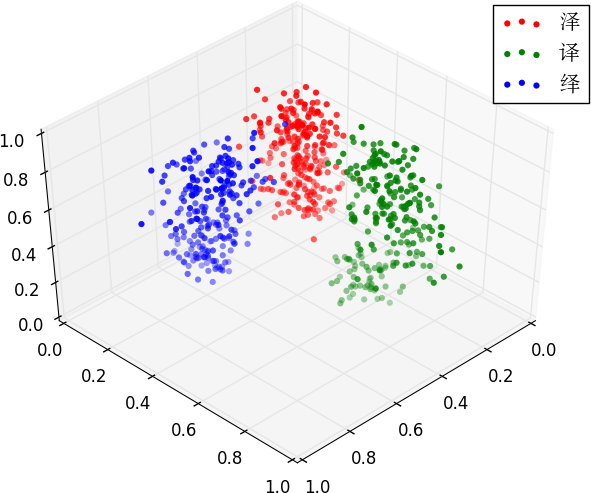} \\
		\includegraphics[width=0.27\linewidth]{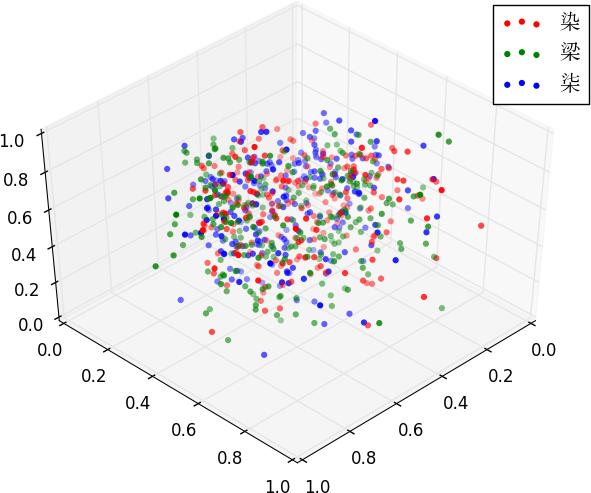} &
		\includegraphics[width=0.27\linewidth]{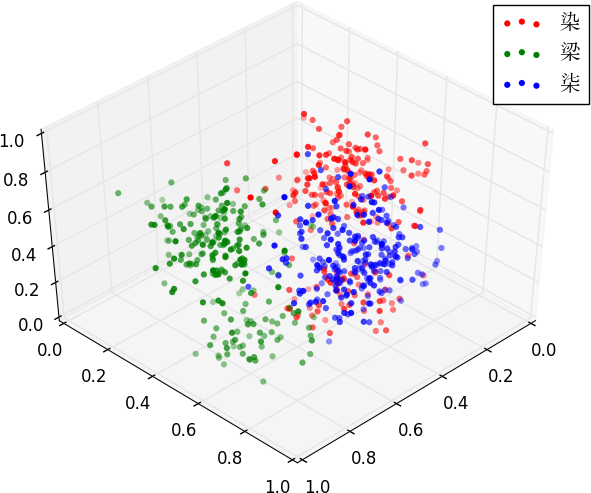} &
		\includegraphics[width=0.27\linewidth]{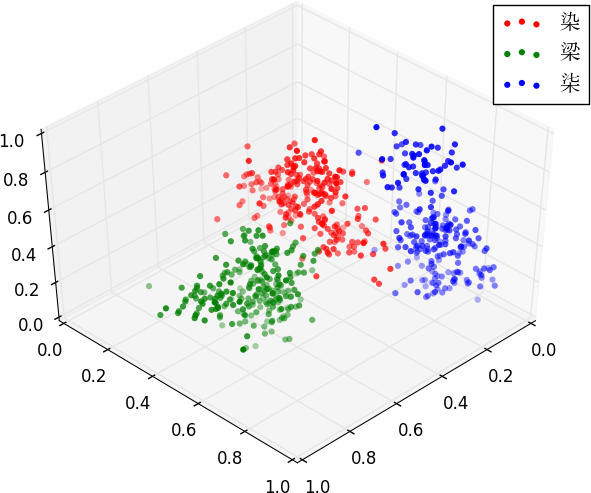} \\
		& & \\
		(a) Original image & (b) Softmax & (c) Template loss
	\end{tabular}
	\caption{Comparison of original image, softmax loss and our template  
		loss. Characters with similar structure can be clearly separated with our proposed loss function.}
	\label{fig:teaser2}
\end{figure*}
The technical advent of computational power in recent years has fueled the
development of deep convolution neural networks (CNNs) which have surpassed 
traditional machine learning in various problem domains. In particular, 
Ciresan et al.~\cite{mcdnn} first proposed the multi-column deep neural network 
(MCDNN), which was composed of several CNNs, to address the HCCR problem. Since 
then, the relevant state-of-the-art performance has been progressively improved.  
In the mean time, CNN is also evolving as well, with numerous new and effective 
network architectures and loss functions proposed to solve various problems.

\begin{figure}[t]
	\centering
	\begin{tabular}{@{\hspace{0mm}}c@{\hspace{4mm}}c@{\hspace{4mm}}c}
		\includegraphics[width=0.12\linewidth]{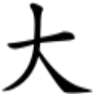} &
		\includegraphics[width=0.12\linewidth]{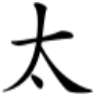} &
		\includegraphics[width=0.12\linewidth]{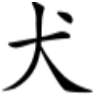} \\
		\includegraphics[width=0.12\linewidth]{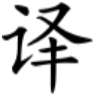} &
		\includegraphics[width=0.12\linewidth]{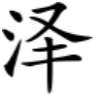} &
		\includegraphics[width=0.12\linewidth]{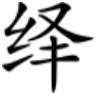}
	\end{tabular}
	\caption{Ambiguous characters. Characters in the first row only differ in one stroke while those in the  second row share the exactly same radical.}
	\label{fig:cha1}
\end{figure}
\begin{figure}[t]
	\centering
	\begin{tabular}{@{\hspace{0mm}}c@{\hspace{10mm}}c}
		\includegraphics[width=0.3\linewidth]{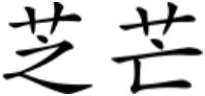} &
		\includegraphics[width=0.27\linewidth]{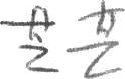} \\
		& \\
		(a) Print & (b) Handwritten
	\end{tabular}
	\caption{Comparison of print and handwritten text. Stroke connection in handwritten 
		text may cause confusion in recognition.}
	\label{fig:cha2}
\end{figure}

To improve classification performance on large scale categories dataset, researchers 
have proposed various loss functions which have achieved impressive results 
in face recognition \cite{liu2016large,liu2017sphereface,AMloss,cosface,arcface}. 
In summary, these loss functions are different variants of softmax loss. 
The intuition is to add a margin on the decision boundary to increase the distance 
between different categories. Nonetheless all previous methods expand a fixed decision margin in 
various manifolds. However, the Chinese character system is highly complicated. 
We observe two main difficulties: (1) many characters are 
similar, or partially similar in structure.  Some print characters hold the common skeleton, 
e.g., the difference of three examples in the first row of Figure~\ref{fig:cha1} 
consist of one stroke only. On the other hand, most characters are composed of 
a number of radicals.  Different characters may contain the same radical, and 
most of characters are partially 
similar, e.g., the second row of Figure~\ref{fig:cha1}; (2) Handwritten strokes can easily lead to 
recognition confusion. Comparing to printed text, handwriting is usually distorted. Moreover,
cursive handwriting produces stroke connections which can obscure the original structure
of the character. For example, in Figure~\ref{fig:cha2}, the vertical stroke makes it 
much more difficult in distinguishing the two handwritten characters. 

To handle the first challenge, we use print characters as templates of handwritten text.
From character templates, we propose a measurement prior for template 
similarity. First we choose a set of image features to produce a 
measurement of print character similarity individually, followed by integrating 
them through dimension reduction to generate the final distance. 
Then, we set a prior margin between each class based on the estimated distance. 
We choose 8 kinds of standard Chinese font as templates and compute a mean feature over them.
The final distance can sufficiently measure the similarity between two given 
templates. 

To address the second challenge, we introduce an instance level loss function. 
Each handwritten character can be treated as an instance of template, with stroke 
variance introduced. we observe that 
the lack of clarity in handwriting introduces a number of difficult examples, 
which are far from class center. However 
there are still a lot of clearly written samples that can be easily distinguished by the classifier.
In other words,  there exists an imbalance between hard and easy samples. To increase the attention 
paid to the outliers, we propose 
our adaptive margin to dynamically fit the training loss to bias hard samples. 
The intuition is to add a moderating factor that is controlled by the estimated 
probability for the ground-truth class.  Specifically, if the sample is far from 
class center, it will be given a large margin to other classes, thus increasing its contribution 
in the final loss function, and vice versa. With this re-balancing moderation, 
badly written characters will be given more attention. Outliers will be contracted to 
the class center, reducing category variance. We will 
demonstrate the effectiveness of our adaptive margin in the experiment section. 

In summary, in this paper we propose two new types of loss function to deal 
with HCCR difficulties. Our contribution includes: 1) a new prior 
margin to handle character template similarity; 2) a new adaptive margin
to reduce category instance variance; 3) an elaborate network architecture 
to accomplish HCCR task.  Extensive experiments demonstrate that our networks 
are able to achieve state-of-the-art performance.


\section{Related Work}

\begin{figure}	
	\centering
	\includegraphics[width=1.0\linewidth]{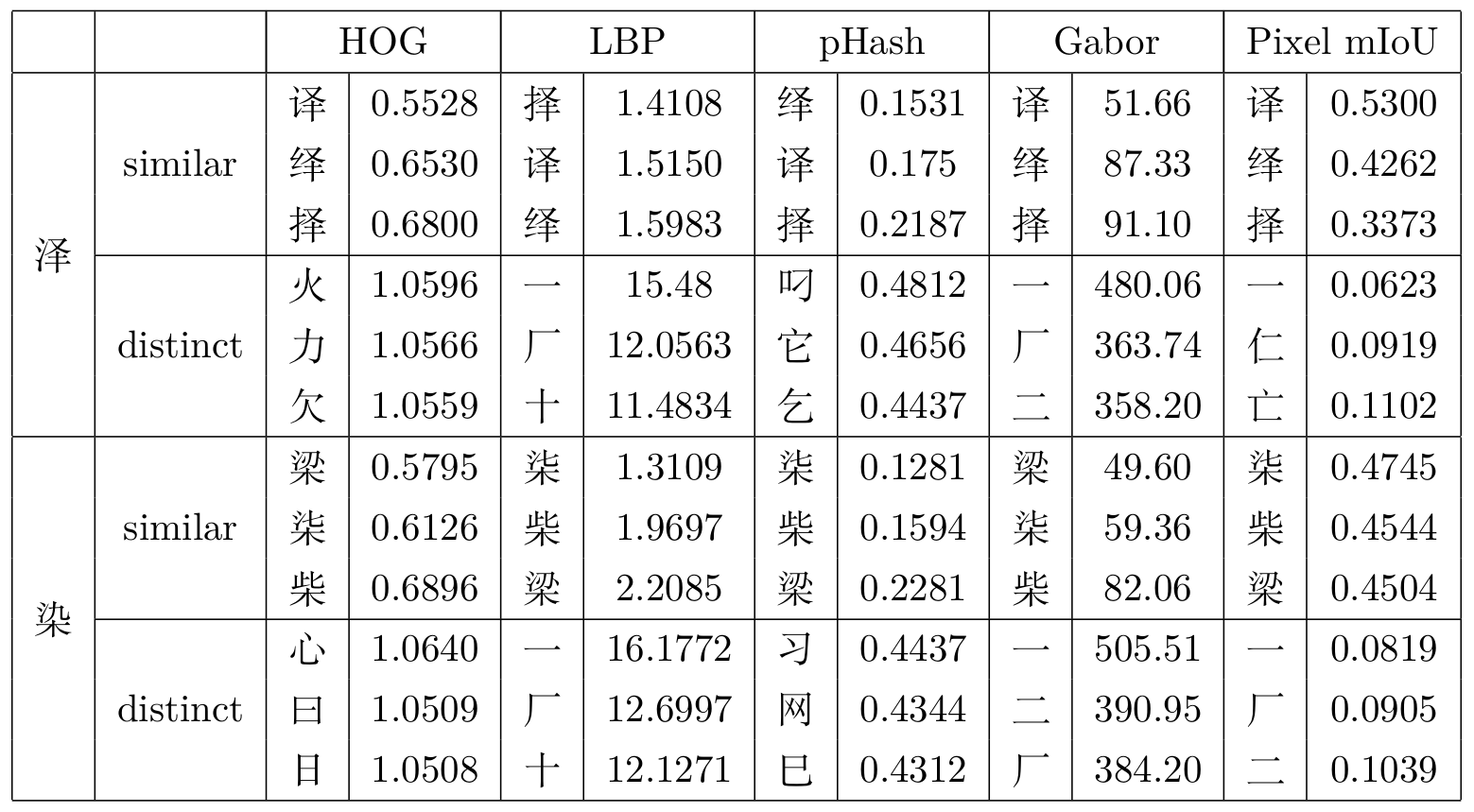}
	\caption{Estimated distance of templates.}
	\label{fig:prior}
\end{figure}

\subsection{Offline HCCR}
Handwritten Chinese Character Recognition (HCCR) has been intensively studied since 1980s. Traditional HCCR methods have three procedures: pre-processing, hand-crafted feature extraction, and classification. Recently, significant progress has been achieved by leveraging the great success of deep convolutional neural networks (CNN) in various domains. 
Multi-column deep neural network (MCDNN)~\cite{mcdnn} was the first successful application of CNN to offline HCCR, with the best performance of single network achieving an accuracy of 94.47\%, which was improved to 95.78\% by ensembling 8 networks. Since then, more researchers have shifted their attention from traditional methods to deep CNN models for HCCR. In the offline HCCR competition held by ICDAR in 2013, Yin et al.~\cite{yin2013icdar} won the first place with the recognition rate of 94.77\%. A year later, the accuracy was improved to 96.06\% by using a voting strategy on four ATR-CNNs~\cite{wu2014handwritten}. In 2015, Zhong et al.~\cite{zhong2015high} combined traditional feature extraction methods with GoogLeNet, and obtained an accuracy of 96.35\%, 96.64\%, and 96.74\% with single, four and ten ensemble models, respectively. With these results, accuracy reported in ~\cite{zhong2015high} became the first to surpass human performance. In 2017, Zhong et al.~\cite{zhang2017online} further improved the accuracy to 96.95\% by the combination of traditional DirectMaps and CNNs. Xiao et al.~\cite{xiao2017building} designed a nine-layer CNN model with network pruning and achieved an accuracy of 97.09\%, with only 1/18 size of its original size.

\subsection{Loss Function}
Loss function is analogous to a ruler for deep learning, which is used to measure a given model and guide the training. Therefore, the design of loss function is as important as the model structure design. Loss function is used to evaluate the relationship between the predicted value and the ground truth value. Euclidean loss, hinge loss, softmax loss, etc. are widely used in deep CNN models. Contrastive loss~\cite{hadsell2006dimensionality} increases the inter-class and reduces the intra-class variance to get better deep features. Similar to contrastive loss~\cite{hadsell2006dimensionality}, triplet loss~\cite{schroff2015facenet} introduces triple sample pairs \{x: input, x1: positive, x2: negative\} to increase Euclidean margin for better feature embedding. Since contrastive loss and triplet loss need to use pair label information, both of them require a careful design of pair selection procedure. Center loss~\cite{wen2016discriminative} was proposed to learn centers for deep features of each sample. The distribution of these deep features is uniform within the same class, which helps to reduce intra-class variance. Large margin softmax (L-softmax)~\cite{liu2016large} adds cosine distance constrain to each identity, with the aim of enhancing the inter-class compactness and inter-class separability. Deep CNN models with angular softmax (A-softmax)~\cite{liu2017sphereface} can learn more discriminative features by normalizing the weights based on L-softmax.

Large margin cosine loss~\cite{cosface}, Additive Margin Softmax (AM-softmax)~\cite{AMloss} and Additive angular margin loss~\cite{arcface} all transform the original softmax loss function into a cosine loss style by normalizing the deep features and the weights. The normalization eliminates the variance in radial direction (compared to L-Softmax) and a cosine based margin is introduced. This margin was used to further maximize the decision boundary of the learned features in the angular space. Wan~\cite{LGM2018} proposed Large-margin Gaussian Mixture (LGM) loss, assuming that deep feature satisfied Gaussian mixture distribution and a certain classification margin at the same time. Lin~\cite{focalloss} proposed focal loss to make the model more focused on difficult samples by reducing the weight of the easily classified samples.

\section{Template-Instance Loss Functions}
We propose two novel loss function, namely template and instance loss. For 
the $i^{th}$ category, we derive our loss from additive margin~\cite{AMloss} due to its simplicity:
	\begin{equation}
	L_i = -\log\frac{e^{s(\cos\theta_{y_i}-m)}}{e^{s(\cos\theta_{y_i}-m)}+\sum_{j=1,j\neq y_i}^{n} e^{s\cos\theta_j}}
	\label{eq:additive}
	\end{equation}
where $y_i$ is the ground-truth label, $s$ is a scale factor and $m$ is the embedded margin. 

\subsection{Template Loss}
There exists intrinsic similarity among different templates of Chinese characters due to 
the unique structure of the Chinese character system.  The Chinese dictionary provides isomorphic characters 
as reference. Typically these look-like characters are searched by humans, and the 
relevant similarity cannot be quantified. We seek to compare all commonly used characters templates to 
calculate an template affinity matrix as the prior margin.

There is no existing rule to define Chinese characters similarity. To 
quantify how two templates look like, we assemble a number of classical vision 
features as feature pool, including pixel-wise 
mIoU, pHash, SIFT, Hog and Gabor feature. To sidestep the irregularity of handwriting, 
we utilize gray-scale images of multiple standard fonts, such as SimSun, 
SimHei, FangSong, KaiTi and so on, as templates and compute features upon them. The average 
template feature distance is computed 
as the mean over all the fonts. See Figure~\ref{fig:prior} for demonstration. 


However, different features are unrelated and exist in different spaces. We need to 
integrate them to produce our template affinity representation. Hence a proper normalization 
is needed. A naive solution is to linearly rescale each individual feature similarity 
to the range [0,1]. However some distance has unlimited upper bound which leads to scale imbalance. 
Instead, we perform a dimension reduction to generalized normalized score like  \cite{masuda2017random}. 

Given $K$ features and $N$ templates, we define directional similarity $N \times K$ matrix $\mathbf{S}^i$ , 
which encodes the similarity between template $i$ and others over all features. 
Element $s^i_{k,j}$ represents similarity of feature $k$ between template $i$ and $j$. 
Since $\mathbf{S}^i$ is not symmetric, we compute its SVD decomposition
	\begin{equation}
	\mathbf{S}^i=\mathbf{U}^i\mathbf{S}^i{\mathbf{V}^i}^T
	\end{equation}
and the normalized ($N$ dimensions ) similarity vector $\mathbf{\hat{h}}^i$ is defined as the 
column of $\mathbf{V}^i$ corresponding to the maximum singular value. Similar to 
eigenvector, the elements of principal column of $\mathbf{V}^i$ are a measures 
of centrality, and are proportional to the amount of time that an infinite-length 
random walk on $\mathbf{S}^i$ will spend \cite{masuda2017random} to measure the distance of two templates. 
We stack the $N$ normalized vector for all templates to obtain a matrix $\mathbf{H}$ and 
the affinity matrix is expressed as 
	\begin{equation}
	\mathbf{A} = \frac{\mathbf{H}+\mathbf{H^T}}{2}
	\end{equation}
which will be rescaled to normalize the diagonal elements. $a_{ij}$ indicate the similarity of two templates. 

With the affinity matrix we define our prior margin between template $i$ and $j$ 
as the softmax probability:
	\begin{equation}
	m^P(i,j)= \left\{ \begin{aligned}
	&\frac{e^{a_{ij}}}{\sum_j e^{a_{ij}}}, & j\neq i \\ &0, & j=i 
	\end{aligned} \right.
	\end{equation}
where $a_{ij}$ is the $(i,j)$ element of $\mathbf{A}$. We embed the margin into ~Equation \ref{eq:additive}:
	\begin{equation}
	L_i=-\log\frac{e^{z_{y_i}}}{\sum_{j=1}^{n}e^{z_j+\beta m^P_{ij}}}
	\end{equation}
where $\beta$ is a scaling factor. After integrating the prior margin, highly similar template 
pair needs a larger decision boundary to be distinguished. Note that the prior margin is 
fixed for the whole category. Template loss works on class level, affecting 
the class center and scale. As for the hard samples or outliers, we need another loss function to 
deal with individually. 

\label{HCCR Networks}
\begin{figure*}	
	\centering
	\includegraphics[width=0.75\linewidth]{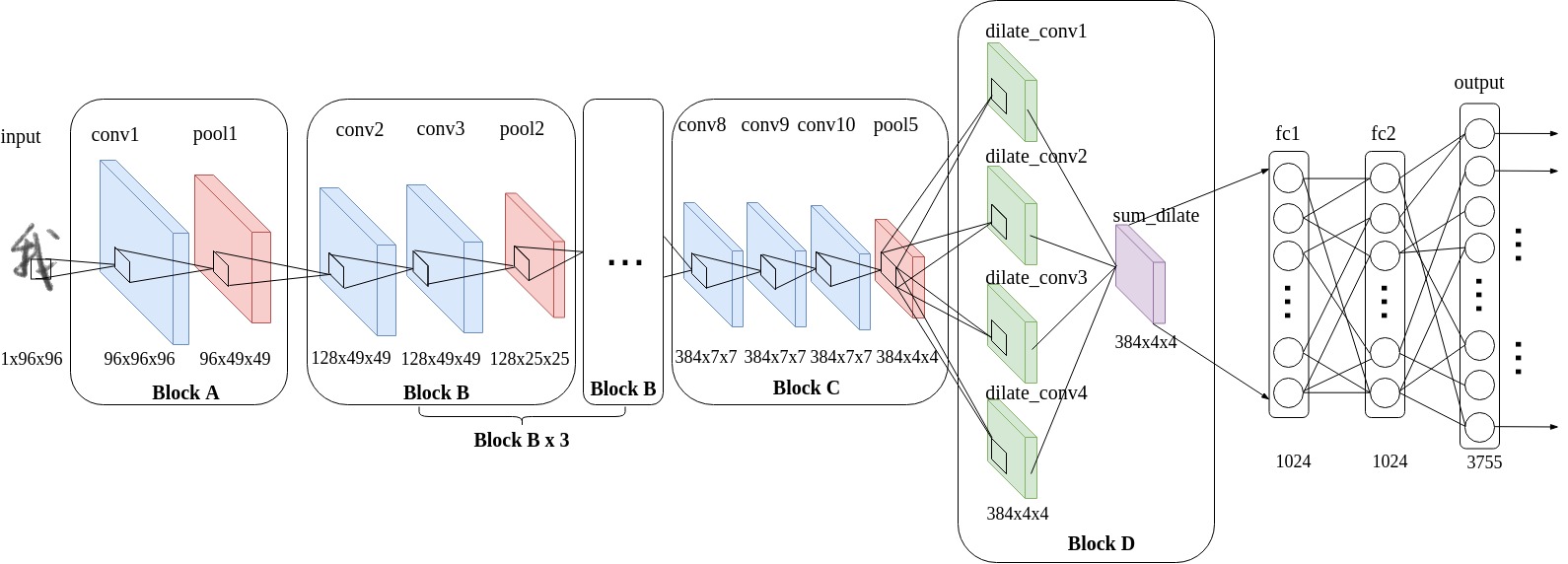}
	\caption{Network architecture.}
	\label{architecture}
\end{figure*}

\subsection{Adaptive Margin}\label{sec:adap}

Recent methods~\cite{liu2016large,liu2017sphereface,AMloss,cosface,arcface} 
all expand a fixed decision margin in 
various manifolds. These margins constrain trained features to be compact 
around class center, including both hard and easy samples. However, 
according to~\cite{focalloss}, in the detection process there exist a large number
of background samples which are easy to classify. These easy samples 
can overwhelm hard positive ones and dominate the final loss.  Similar situations
occur for Chinese characters, because 
well-written instance could be easily distinguished, while badly-written ones 
are hard to be recognized. Since common radicals occur among distinct Chinese characters, 
stroke confusion raises classification difficulty.
Therefore, it is of significant importance to adaptively impose the margin according to 
classification difficulty of each instance. We propose instance loss, which is 
achieved by adaptive margin based on the softmax loss. To evaluate the difficulty of input 
instance, we use the estimated probability for the ground-truth label $y_i$:
	\begin{equation}
	p_i = \frac{e^{z_{y_i}}}{\sum_j e^{z_j + \beta m^P_{ij}}}
	\label{eq:p}
	\end{equation}
Then we define our adaptive margin as 
	\begin{equation}
	m^A_i = \alpha (1-p_i)^\gamma
	\label{eq:m}
	\end{equation}
where $\alpha$ is the upper-bound of margin. Our margin has two desirable 
properties: (1) When an instance is far from class center and misclassified, 
$p$ should be small and the margin is maximized to its upper-bound $\alpha$, thus leading 
to larger penalty for the misclassification. 
As $p \to 1$, the margin goes to $0$ and the loss for well-classified samples 
will be close to the original softmax loss; (2) The parameter $\gamma$ 
smoothly adjusts the margin scale between easy and hard samples. 
When $\gamma=0$, $m$  is the same as additive margin, and as $\gamma$ 
increased the effect of margin adaptability is also enhanced. 
Since $m_i^A$ depends on $p_i$, which is the estimated probability of individual 
sample, the adaptive margin works on instance level, increasing the punishment 
of difficult samples or outliers.

Putting Equations \ref{eq:p} and~\ref{eq:m} into \ref{eq:additive} we get:
\begin{footnotesize}
	\begin{align}
	L_i &=-\log \frac{e^{s(\cos\theta_{y_i}-m^A_i)}}{e^{s(\cos\theta_{y_i}- m^A_i)
		}+\sum_{j=1, j\neq y_i}^{n}e^{s\cos\theta_j + \beta m^P_{ij}}} \nonumber \\
	&=-\log \frac{e^{-sm^A_i}}{\frac{\sum_{j}^{n}e^{s\cos\theta_j + \beta m^P_{ij}}}{e^{s\cos\theta_{y_i}}}+e^{-sm^A_i}-1} \nonumber \\
	&=-\log \frac{e^{-sm^A_i}}{\frac{1}{p}+e^{-sm^A_i}-1} \nonumber \\
	&=-\log \frac{1}{(\frac{1}{p_i}-1)e^{s\alpha (1-p_i)^\gamma}+1}
	\end{align}
\end{footnotesize}
In practice the $\alpha$ is not a fixed parameter. It increases with 
iteration number rising in a sigmoid style, i.e., at the beginning of training 
the loss is approximately the same with softmax. With the training proceeding, 
the margin influence will gradually increase, thus weakening contribution 
of easy samples. We will further discuss the setting of $\alpha$ in experiment 
section.

\section{HCCR Networks}

The architecture of our designed network is shown in Figure~\ref{architecture}. The input image
is resized to $96\times96$ which slightly increases performance. Except for converting the RGB image into gray format and subtracting the mean gray value, we do not perform other preprocessing. The resized images are then passed through a stack of blocks and fully connected layers. 

In this paper, we design four kinds of blocks, namely, Block A to Block D. Block A consists of one Conv-BN-PReLU module, followed by a max pooling layer over a $3\times 3$ window with stride of 2 and one pixel padding. The only difference between Block A, Block B, and Block C is the number of Conv-PReLU-BN modules. In Block B and Block C, the number of Conv-BN-PReLU modules is two and three, respectively. Convolutional kernels in Block A to C are $3\times 3$, with padding size of 1 and stride of 1. Block D is a hybrid
dilated convolution (HDC) module~\cite{AAAI2018}, which includes a multi-dilated convolution layer and a pixelwise summation of the multi-dilated convolutional feature maps. After dilation, the dilated convolutional kernel size is changed from $k\times k$ into $r\times\left(k-1\right)+1$, where $k\times k$ is the original convolutional kernel size, and $r$ is the dilation rate. In order to retain the size for summation, the padding pixels are computed in accordance with the dilation rate of the kernel. In this paper, after the last pooling layer, the size of feature maps is $4\times 4$. The kernel sizes for the 4-channels dilation convolutional layers are all $3\times 3$, with dilation rates and padding pixels ranging from 1 to 4. These multi-dilated convolutional feature maps are summed up before sending into the subsequence fully connected fc1, fc2 and output layers. Fc1 and fc2 contains 1024 neurons, equipped with batch normalization (BN)~\cite{ioffe2015batch} and parametric rectified linear unit (PReLU)~\cite{he2015delving}. Features in fc2 layer are applied L2 normalization before sent into the last output layer, which has 3755 neurons to carry out the final Chinese character recognition task. 


The overall architecture is illustrated in \ref{architecture}, which contains 10 Conv-BN-PReLU modules, 5 pooling layers, 1 multi-dilated convolutional, and 3 fully connected layers. As a result, our network architecture is designed as a 14-layer network (only accounting for the convolutional and fully connected layers). We refer to this network as HCCR14layer.

\begin{figure*}[t]
	\centering
	\begin{tabular}{@{\hspace{0mm}}c@{\hspace{3mm}}c@{\hspace{3mm}}c@{\hspace{3mm}}c}
		\includegraphics[width=0.207\linewidth]{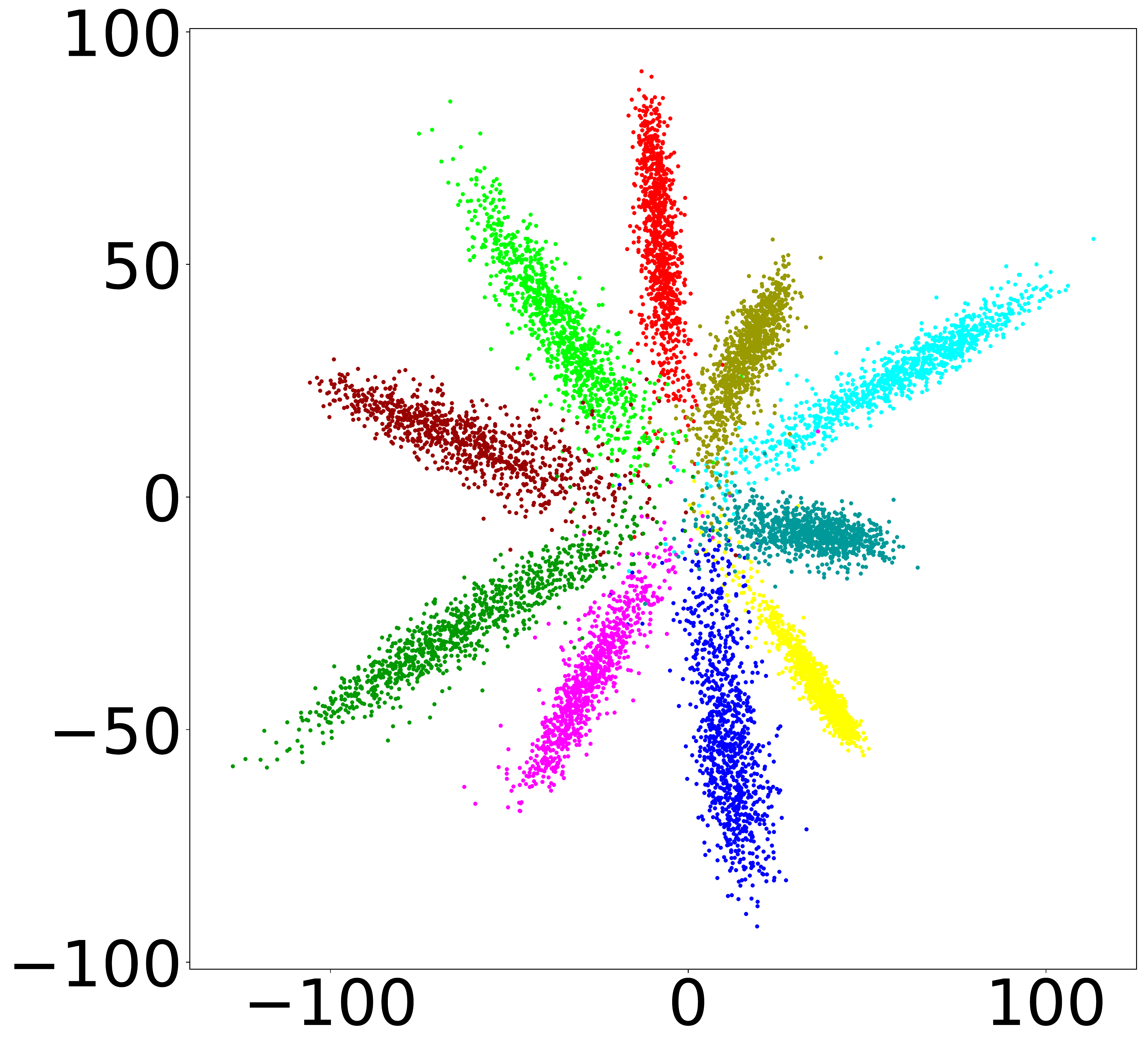} &
		\includegraphics[width=0.2\linewidth]{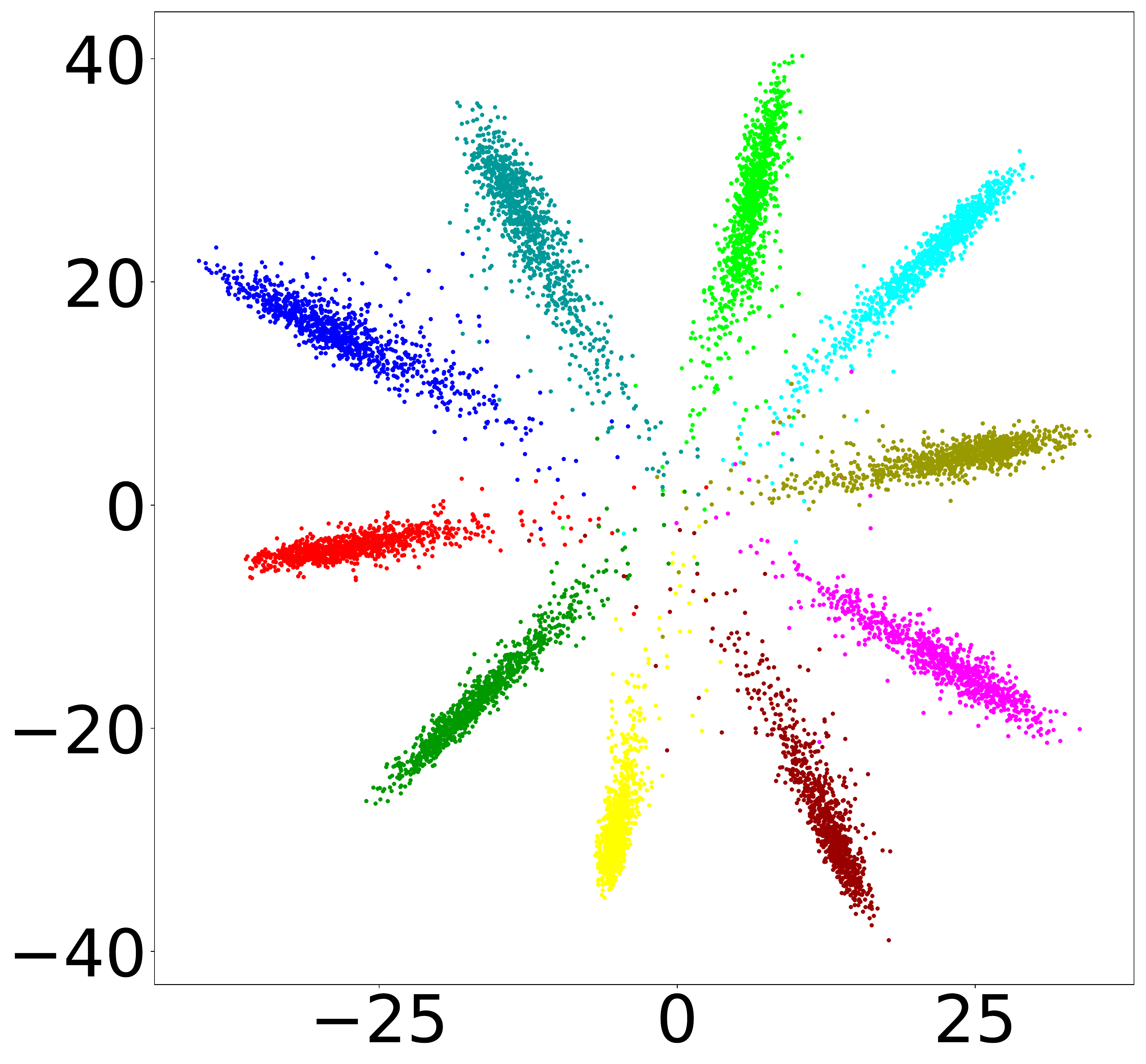} &
		\includegraphics[width=0.203\linewidth]{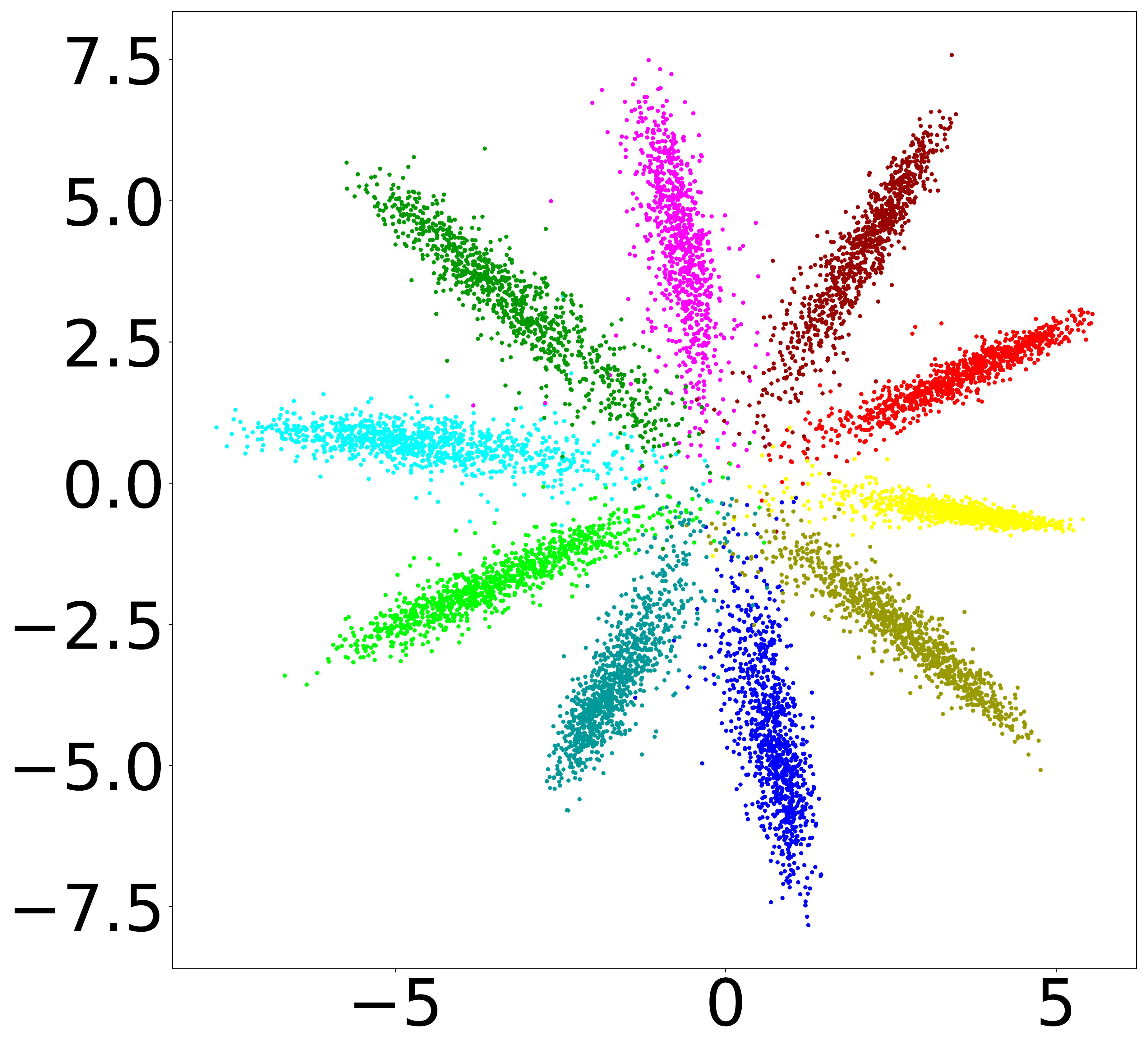} &
		\includegraphics[width=0.2\linewidth]{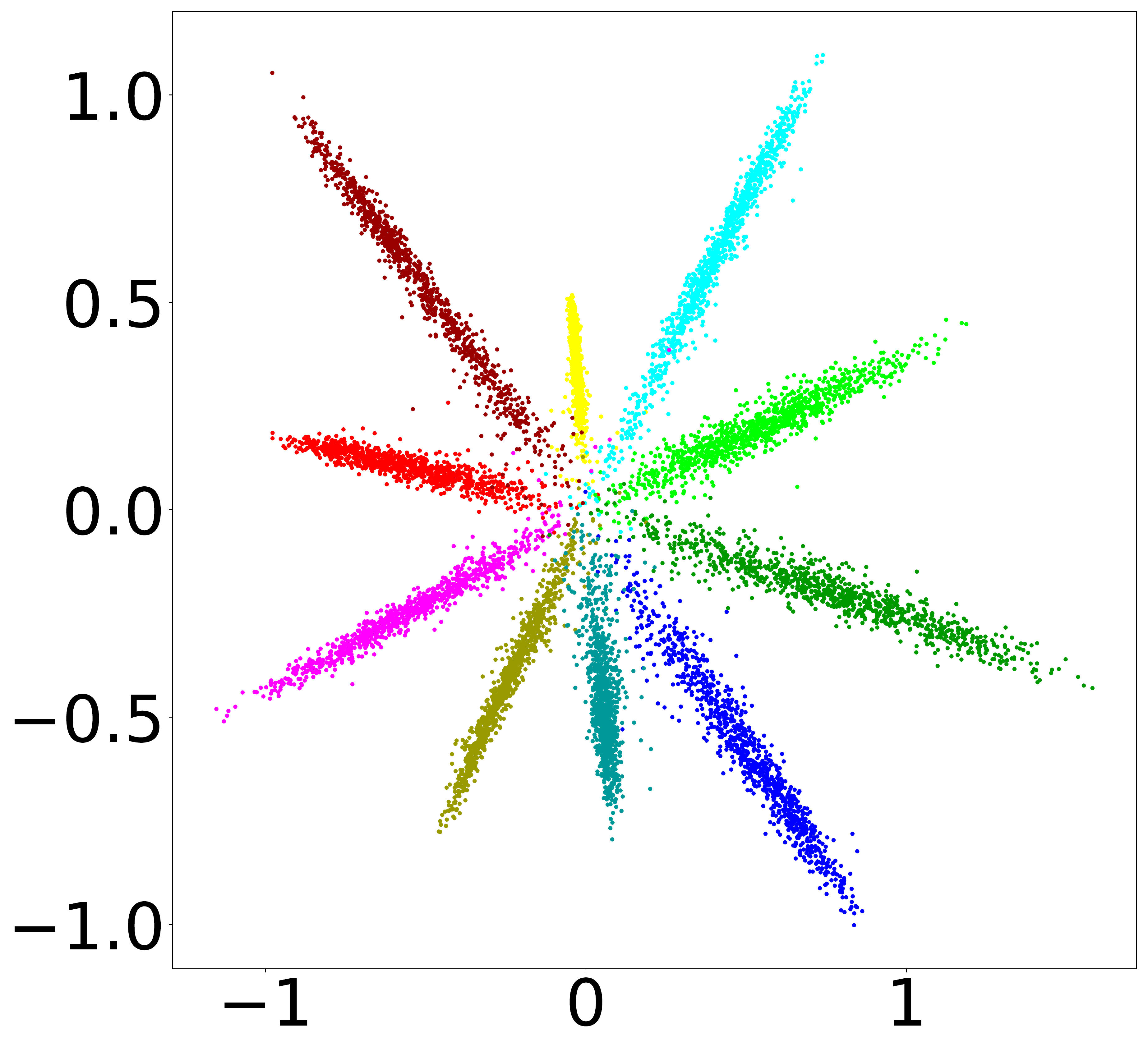} \\
		(a) & (b) & (c) & (d) \\
		\includegraphics[width=0.2\linewidth]{fig/loss/ours/01} &		
		\includegraphics[width=0.2\linewidth]{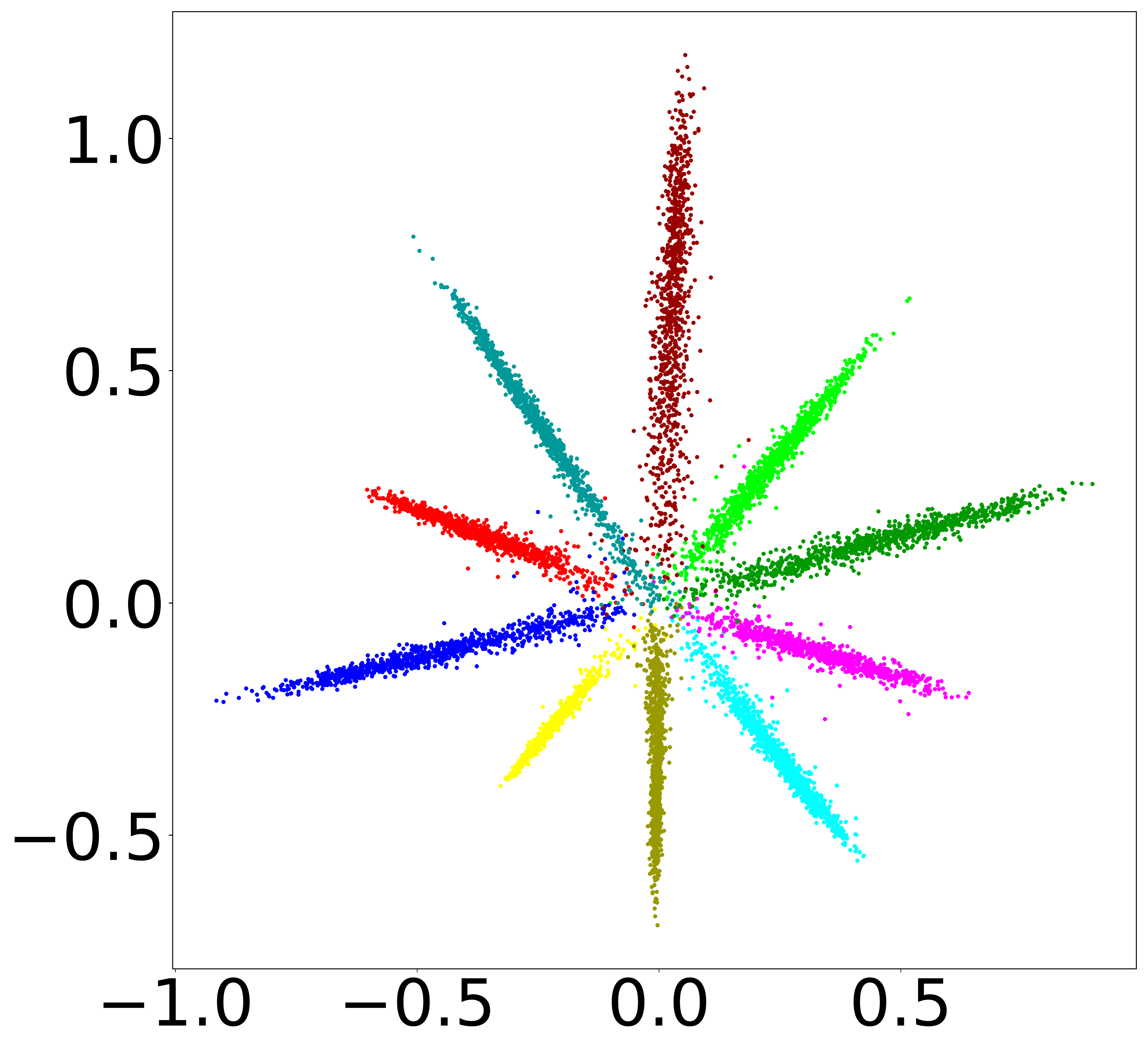} &
		\includegraphics[width=0.2\linewidth]{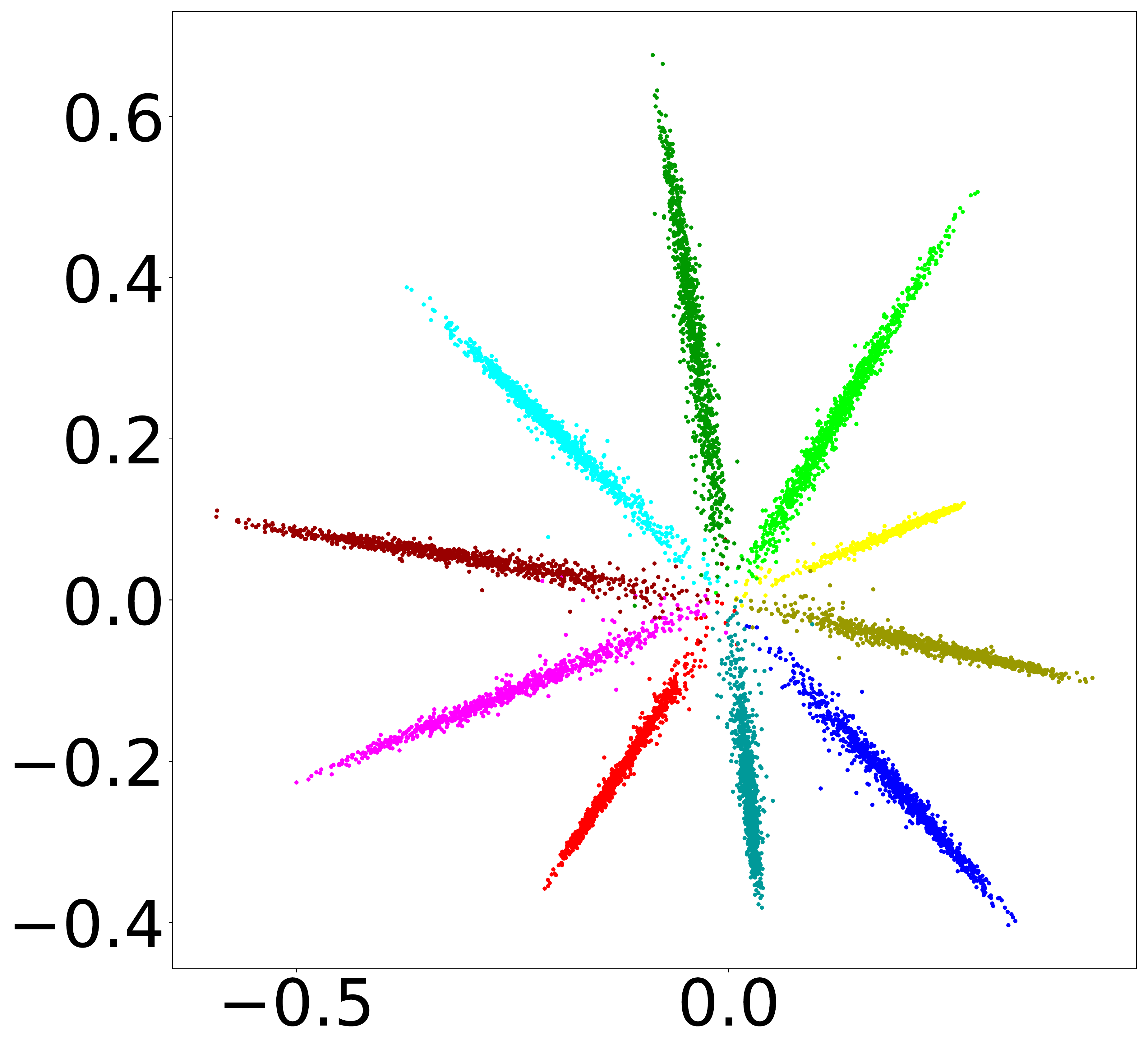} &
		\includegraphics[width=0.2\linewidth]{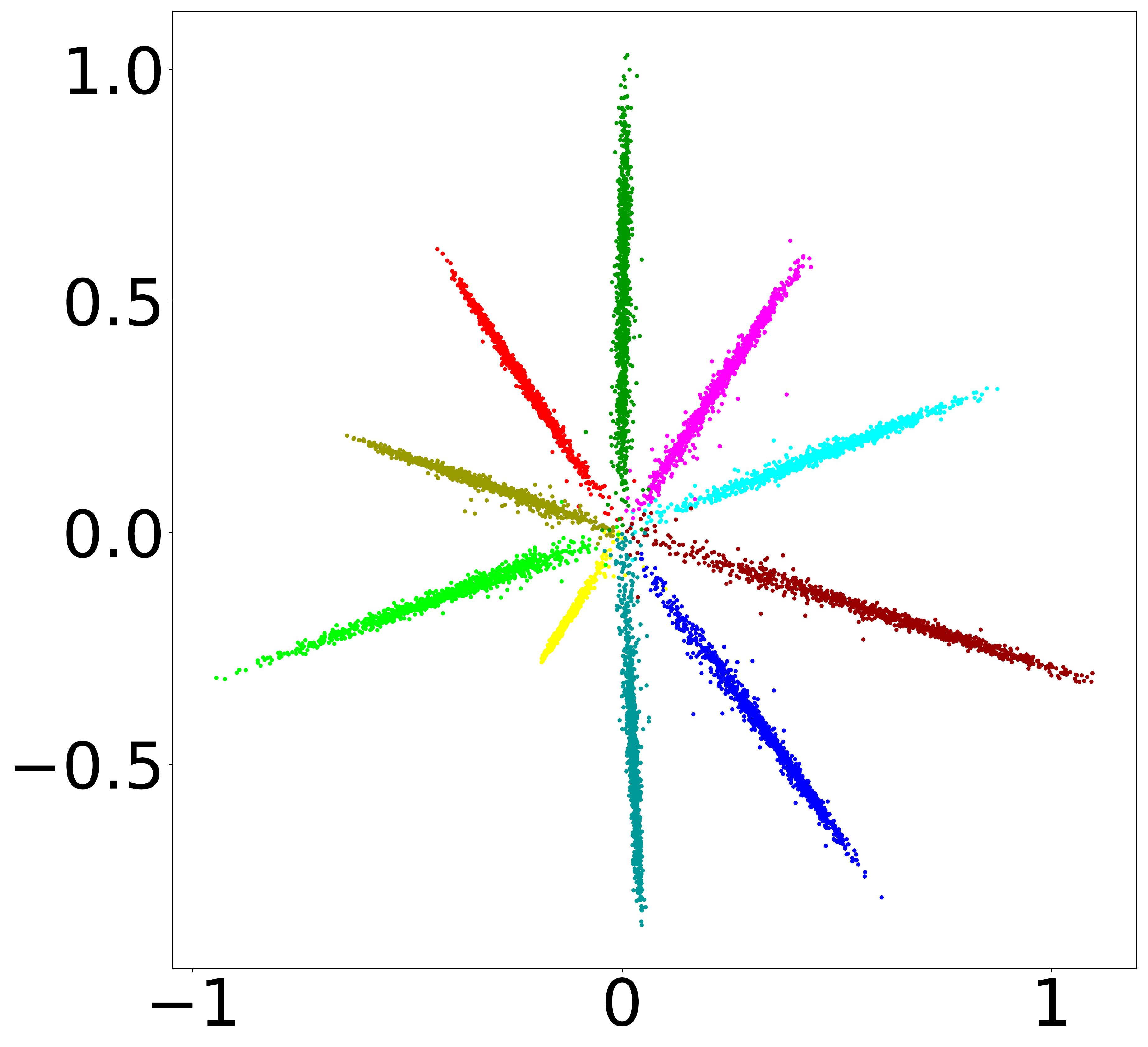} \\
		(e) & (f) & (g) & (h) \\
	\end{tabular}
	\caption{Visualization of two-dimensional feature embeddings on MNIST training set. First row: (a) Softmax loss. 
		(b) Large-margin softmax loss \cite{liu2016large}. (c) Additive margin loss \cite{AMloss}. (d) Adaptive 
		margin loss. Second row: (e-h) Adaptive margin with $\alpha=0.1,0.2,0.3,0.4$. Margin effect expands with $\alpha$ increasing.}
	\label{fig:margin}
\end{figure*}

The main difference between our proposed HCCR14layer and other previous models for offline HCCR is that HCCR14layer involves PReLU, BN, and HDC layer. In our proposed network, all convolutional layers in Block A, B, C, and two fully-connected layers (fc1 and fc2) are followed by BN and PReLU layer. The BN layer was proposed by Ioffe et al.~\cite{ioffe2015batch}, which can not only efficiently speed up network convergence, but also effectively avoid the problem of vanishing gradients during training. The PReLU~\cite{he2015delving}, with a parametric constrain based on ReLU~\cite{nair2010rectified}, was shown to be able to handle vanishing and exploding gradient. Since PReLU~\cite{he2015delving} contains only a few parameters, the network can still easily converge to the optimal state, while risking a bit to overfitting. 

The template and instance loss functions are combined as the final loss. 
Our proposed loss will rebalance the contribution of training samples according to template similarity 
and instance difficulty, improving model performance. Although the proposed HCCR14Layer model employs quite simple layers, it yields state-of-the-art performance for offline HCCR by cooperating with our novel template-instance loss.

\section{Experiments}

We conducted extensive experiments to verify the effectiveness of our novel 
loss. 
We also present experimental results of our HCCR networks and 
perform comparison with state-of-the-art methods.

We used the data obtained from offline CASIA-HWDB1.0 (HWDB1.0) and CASIA-HWDB1.1 (HWDB1.1)~\cite{liu2011casia} for training, and ICDAR-2013~\cite{yin2013icdar} competition data for testing, which were provided by the Institute of Automation of the Chinese Academy of Sciences. The detailed information about these datasets are listed in Table~\ref{table:dataset}. The training set contains 2,678,424 samples written by 720 people, and 3,755 characters in the GB2312-80 level-1 set. The testing dataset contains 224,419 samples written by another 60 people. Before training, we merged HWDB1.0 and HWDB1.1 to get HWDB1.0+1.1 with respect to HWDB1.1's class labels. We have also made sure that the merged dataset HWDB1.0+1.1 and ICDAR-2013 have the same class label.

\subsection{Adaptive and Prior Margin}\label{exp_adapt}

\begin{table}[t]
	\centering
	\begin{tabular}{r|r|r|r|r|r}
		\hline
		$\alpha_{max}$ & 0 (softmax) & 0.01 & 0.05 & 0.1 & 0.2 \\
		\hline
		Accuracy (\%) & 94.49 & 95.32 & 95.59 & \textbf{95.75} & 95.66 \\
		\hline
	\end{tabular}
	\caption{Performance of instance loss with different setting. Note that $\alpha_{max}=0$
		is original softmax}
	\label{table:alpha}
\end{table}

In Figure~\ref{fig:margin} we perform a toy experiment to better visualize 
the margin effect on features. We extract features on MNIST dataset using 
other losses and our proposed adaptive margin. In this experiment we fix 
$\alpha$ for simplicity. We show various results with different $\alpha$s for 
comparison. As shown in the figure, the first row displays the comparison 
of the original softmax, large margin softmax (LM), additive margin (AM) and our adaptive 
margin (AD). We can observe that our adaptive 
margin shows strong margin effect. As $\alpha$ increases, 
the angular margin between classes are amplified, as shown in the second row.

\begin{table*}	
	\centering
	\begin{tabular}{l|c|c|c}
		\hline
		Layers &  A & B & C \\
		\hline
		Bloak A  
		&  $\left[ \tabincell{c}{Conv-BN-PReLU \\Pool} \right]$ $\times 3$ 
		& $\left[ \tabincell{c}{Conv-BN-PReLU \\Pool} \right]$ $\times 1$
		& $\left[ \tabincell{c}{Conv-BN-PReLU \\Pool} \right]$ $\times 1$ \\
		\hline
		Bloak B  
		& $\left[ \tabincell{c}{Conv-BN-PReLU \\Conv-BN-PReLU \\Pool}\right]$ $\times 2$ 
		& $\left[ \tabincell{c}{Conv-BN-PReLU \\Conv-BN-PReLU \\Pool}\right]$ $\times 3$
		& $\left[ \tabincell{c}{Conv-BN-PReLU \\Conv-BN-PReLU \\Pool}\right]$ $\times 3$\\
		\hline
		Bloak C  
		& /
		& $\left[ \tabincell{c}{Conv-BN-PReLU \\Conv-BN-PReLU \\Conv-BN-PReLU \\Pool}\right]$ $\times 1$
		& $\left[ \tabincell{c}{Conv-BN-PReLU \\Conv-BN-PReLU \\Conv-BN-PReLU \\Pool}\right]$ $\times 1$  \\
		\hline
		Bloak D  & \multicolumn{3}{|c}{HDC} \\
		\hline
		Fully Connected Layer 
		&  \tabincell{c}{1024FC-BN-PReLU }
		& \tabincell{c}{1024FC-BN-PReLU } & \tabincell{c}{1024FC-BN-PReLU \\1024FC-BN-PReLU} \\
		\hline			
		Classification Layer 
		& \multicolumn{3}{|c}{3755FC} \\
		\hline
	\end{tabular}
	\caption{Configurations for HCCR (shown in columns). The dropout layer is not shown for different dropout ratios.}
	\label{table:HCCR configuration}
\end{table*}

\begin{table}[t]
	\centering
	\begin{tabular}{l|l|l|l}
		\hline
		Dataset & $\#$samples & $\#$class & $\#$writers \\
		\hline
		HWDB1.0  & 1,556,675 & 3,740 & 420 \\
		HWDB1.1  & 1,121,749 & 3,755 & 300\\
		ICDAR-2013  & 224,419 & 3,755 & 60 \\
		\hline
	\end{tabular}
	\caption{Statistics of offline isolated character datasets}
	\label{table:dataset}
\end{table}

\begin{table}[t]
	\small
	\centering
	\begin{tabular}{l|l|l|l|l|l}
		\hline
		Feature & HOG & LBP & pHash & Gabor & Pixel. mIoU \\
		\hline
		Accu (\%) & 87.61 & 79.63 & 81.66 & 82.97 & 85.47 \\
		\hline
	\end{tabular}
	\caption{Verification of human perception and feature similarity.}
	\label{table:feat}
\end{table}

We also demonstrate the performance boost of HCCR by adaptive margin. In this 
experiment we use our HCCR 12 layers as base network. Original softmax and the 
proposed adaptive margin are applied to compare the results, and template loss is 
not integrated. For the training and 
testing data, we choose CASIA HWDB1.1, which includes 897,758 training and 223,991 
testing samples. Input images are resized to $96\times96$ 
to increase performance, similarly done in~\cite{xiao2017building}.

\begin{table*}	
	\centering
	\begin{tabular}{l|l|l|l}
		\hline
		Method & Input & Image Size & Accuracy (\%) \\
		\hline			
		HCCR-Gabor-GoogLeNet \cite{zhong2015high} & Gabor Feature Maps + Gray Image & $9\times 120\times 120$ & 96.35 \\
		\hline			
		DirectMap+ConvNet \cite{zhang2017online}  & Direct Feature Maps & $8\times 32\times 32$ & 96.95\\
		\hline			
		HCCR-CNN9Layer+Skeleton  \cite{AAAI2018}  & Gray Image + Skeleton & $1\times 96\times 96$ & 96.90 \\ 
		\hline
		HCCR-CNN9Layer\cite{xiao2017building}  & Gray Image & $1\times 96\times 96$ & 97.09 \\ 
		\hline
		HCCR-WAP \cite{li2018building} & Gray Image & $1\times 64\times 64$ & 96.91 \\
		\hline
		\hline
		HCCR14Layer+softmax  & Gray Image & $1\times 96\times 96$ & 97.20 \\
		\hline
		HCCR14Layer+additive margin loss  & Gray Image & $1\times 96\times 96$ & 97.31 \\
		\hline
		HCCR14Layer+template-instance loss  & Gray Image &  $1\times 96\times 96$& \textbf{97.45} \\
		\hline
	\end{tabular}
	\caption{Recognition accuracies on ICDAR-2013 dataset of different network configurations. All models are trained on HWDB1.0+HWDB1.1.}
	\label{table:compare}
\end{table*}

\begin{table}	
	\centering
	\begin{tabular}{l|c}
		\hline
		HCCR config. (Table \ref{table:HCCR configuration}) & Accuracy (\%) \\
		\hline			
		A, without fc1 norm  & 96.80  \\
		\hline
		A, fc1 norm  &  96.84  \\
		\hline
		B, without fc1 norm  & 97.37  \\
		\hline		
		B, fc1 norm  & 97.40   \\
		\hline			
		C, without fc2 norm  &  97.40 \\ 
		\hline
		C, fc2 norm  &  \textbf{97.45}  \\
		\hline				
		C(all conv), without fc2 norm  &  97.35  \\
		\hline		
		C(all conv), fc2 norm  &  97.38  \\
		\hline
	\end{tabular}
	\caption{Recognition accuracies on ICDAR-2013 dataset of different network configurations. All models are trained on HWDB1.0+HWDB1.1.}
	\label{table:hccr accuracy}
\end{table}

The factor $\alpha$ in Equation~\ref{eq:m} is a tuning parameter to control the 
strength of margin. Due to the adaptive mechanism, the difficult samples, which is estimated 
by the softmax probability, will be given a large margin to produce a larger contribution 
to the total loss. However, at the beginning of training, if we train from scratch the 
model cannot classify well training samples and produce  inaccurate estimated probabilities.
Thus the margin contribution will be amplified, leading to either not converging, or fast converging to a local 
optimization. To avoid this we moderate $\alpha$ by a sigmoid function: 
	\begin{equation}
	\alpha=\frac{\alpha_{max}}{1+e^{-\rho(\frac{iter}{max\_iter}-0.5)}}
	\end{equation}
At the training starts, $\mathit{iter}=1, \alpha\to0$. At the end, 
$\alpha\to\alpha_{max}$. When $\alpha=0$, the loss is equivalent to the softmax loss. Thus 
the training loss is smoothly decreasing in the beginning phase. 
As the training is progressing, increasing $\alpha$ imposes extra training difficult, 
resulting in the loss bouncing back. In our experiments $\rho$ is set to be 10.
We also compare results with different 
$\alpha_{max}$ settings. As tabulated in Table~\ref{table:alpha}, with adaptive margin the accuracy 
surpass original softmax. Among all settings $\alpha=0.1$ outperforms others. Therefore 
in our following experiments $\alpha$ is set to be 0.1.

On the other hand, to prove the representative ability of the selected features, 
we conduct experiments to verify consistency between human perception and template feature similarity. 
Three persons were asked to randomly pick three templates (A,B,C) from the alphabet. 
Each person give a judgment if template A is more similar to B, or C. If three people have 
disagreement, we accept the majority opinion. Then we compute the similarity of (A,B) and (A,C) 
to compare to human opinion. The test is repeated 500 times. The accuracy of every feature are 
listed in Table~\ref{table:feat}. We can see that selected features are basically consistent with human 
perception, among them HOG and pixel mIoU perform better.


\subsection{HCCR14Layer}

We trained our proposed HCCR14Layer model (Figure~\ref{architecture}) using mini-batch gradient descent with batch size of 128 and momentum of 0.9 for training. The base learning rate was set to 0.1, and was reduced to 1/10 every 70,000 iterations. Training was completed after 360,000 iterations. Finally, the curve tended to reach a steady state with an accuracy of 97.45\% on the testing set.


We also evaluated other network configurations for HCCR in this paper, and all configurations follow the generic design presented in Section~\ref{HCCR Networks}. 
We report the performance of different configurations (outlined in Table~\ref{table:HCCR configuration}) on the testing dataset in Table~\ref{table:hccr accuracy}. From the table we can see that deeper network can get better results, and feature normalization can slightly increase performance. In Table~\ref{table:hccr accuracy}, we also replace the max pooling layer in Block A, B, and C into convolutional layer with kernel size $3\times 3$ and stride of 2. Compared with the last row in Table~\ref{table:hccr accuracy}, we find that our proposed HCCR14Layer (configuration C in Table~\ref{table:HCCR configuration}) achieves better recognition accuracy. The main reason is that pooling layer in HCCR14Layer block modules can not only simplify the complexity of network when computing, but also extract the representative features during feature compression.

In Table~\ref{table:compare}, we compared our proposed network and loss with the other available methods and losses. It is noteworthy that our proposed approach outperforms the other HCCR recognition models, achieving a state-of-the-art recognition rate of 97.45\%.

\section{Conclusion}

Handwritten Chinese character recognition has been a long-standing problem
until recent development of deep CNNs which produced revolutionary improvement to
recognition accuracies.  In this paper we propose template-instance loss function 
to address HCCR: 1) a new prior margin to handle character template similarity; and 
2) a new adaptive margin to rebalance easy and difficult instance.  We propose
and implement a carefully designed network architecture to accomplish HCCR task
using our new loss functions.  Extensive experiments demonstrate that our network
has pushed the frontier of recognition accuracy beyond the existing state-of-the-art 
methods. In the future, we plan to apply our template-instance loss on other classification 
tasks such as object detection and face recognition. We believe that the two-level type 
loss function should generally boost classification performance on large scale dataset.

%
%

{\footnotesize
	\bibliographystyle{ieee}
	\bibliography{egbib}
}

\end{document}